\documentclass[lettersize,journal]{IEEEtran}
\usepackage{amsmath,amsfonts}
\usepackage{algorithmic}
\usepackage{algorithm}
\usepackage{array}
\usepackage[colorlinks,
linkcolor = blue,
anchorcolor = blue,
citecolor = blue,
urlcolor = magenta
]{hyperref}
\usepackage{textcomp}
\usepackage{stfloats}
\usepackage{url}
\usepackage{verbatim}
\usepackage{graphicx}
\usepackage{cite}
\usepackage{enumitem}
\usepackage{multirow}
\usepackage{multicol}
\usepackage{booktabs}
\usepackage{makecell}
\usepackage{bm}
\usepackage{color} 
\usepackage{soul}
\usepackage{mathtools}
\usepackage{textcomp}
\usepackage{gensymb}
\usepackage{enumitem}
\usepackage{amssymb}
\usepackage{pifont}
\usepackage{amsthm}
\usepackage{multirow}
\usepackage{rotating}
\usepackage{adjustbox}
\usepackage[dvipsnames]{xcolor}
\usepackage{bbding}

\usepackage{threeparttable}

\begin{document}

\title{DSFormer: A Dual-Scale Cross-Learning Transformer for Visual Place Recognition}

\author{Haiyang~Jiang\textsuperscript{1},
  Songhao~Piao\textsuperscript{1*},
  Chao~Gao\textsuperscript{2*},
  Lei~Yu\textsuperscript{3},
  Liguo~Chen\textsuperscript{4},

\thanks{\textsuperscript{*}Corresponding author: S. Piao. and C. Gao. }
\thanks{\textsuperscript{1} H. Jiang and S. Piao are with the Multi-agent Robot Research Center, Department of Faculty of Computing, Harbin Institute of Technology, Harbin, 150001 China. E-mail: jianghy.hgd@stu.hit.edu.cn, piaosh@hit.edu.cn.

\textsuperscript{2} C. Gao is with the Institute for AI Industry Research (AIR), Tsinghua University, BeiJing, 100084, China. Email: chao.gao@cantab.net

\textsuperscript{3} L. Yu is with the School of Artificial Intelligence, Wuhan University, Wuhan, 430072, China. E-mail: ly.wd@whu.edu.cn.

\textsuperscript{4} L. Chen is with the Soochow University, Soochow, 215031, China. E-mail: chenliguo@suda.edu.cn.

Sponsored by Xinchen Qihang Inc.}
}

\maketitle

\begin{abstract} 
Visual Place Recognition (VPR) is crucial for robust mobile robot localization, yet it faces significant challenges in maintaining reliable performance under varying environmental conditions and viewpoints. To address this, we propose a novel framework that integrates Dual-Scale-Former (DSFormer), a Transformer-based cross-learning module, with an innovative block clustering strategy. DSFormer enhances feature representation by enabling bidirectional information transfer between dual-scale features extracted from the final two CNN layers, capturing both semantic richness and spatial details through self-attention for long-range dependencies within each scale and shared cross-attention for cross-scale learning. Complementing this, our block clustering strategy repartitions the widely used San Francisco eXtra Large (SF-XL) training dataset from multiple distinct perspectives, optimizing data organization to further bolster robustness against viewpoint variations. Together, these innovations not only yield a robust global embedding adaptable to environmental changes but also reduce the required training data volume by approximately 30\% compared to previous partitioning methods. Comprehensive experiments demonstrate that our approach achieves state-of-the-art performance across most benchmark datasets, surpassing advanced reranking methods like DELG, Patch-NetVLAD, TransVPR, and R2Former as a global retrieval solution using 512-dim global descriptors, while significantly improving computational efficiency. The source code will be released at
\href{https://github.com/aurorawhisper/dsformer.git}{https://github.com/aurorawhisper/dsformer.git}.

\end{abstract}


\section{Introduction}
\label{sec:1}

\IEEEPARstart{V}{PR} serves as a fundamental capability in robotic systems, enabling robots to coarsely locate themselves within an environment by matching visual inputs to a pre-existing geo-tagged database, which is critical for robotics to large-scale geolocation tasks. However, achieving robust and efficient recognition in dynamic real-world, several significant challenges remain. Challenges such as conditional variations ( {\it e.g.}, Illuminations, weather, seasonal changes, long-term, et al. ), viewpoint changes, and perceptual aliasing~\cite{lowry2015visual} (where distinct locations exhibit repetitive or structurally similar patterns.), degrade feature consistency and distinctiveness. Furthermore, practical robotic systems often demand real-time performance under stringent memory and computational constraints, further exacerbating these challenges.

Conventional VPR methods~\cite{khaliq2022multires, sarlin2019coarse, RARS19, kim2017learned} typically rely on CNN backbones ( {\it e.g.}, ResNet~\cite{he2016deep} and VGG~\cite{simonyan2014very}, et al. ) pretrained on ImageNet~\cite{deng2009imagenet} to extract local features. These features are subsequently aggregated into compact descriptors using techniques like GeM~\cite{radenovic2018fine} or NetVLAD~\cite{arandjelovic2016netvlad}, and the resulting models are fine-tuned on VPR-specific datasets, such as MSLS~\cite{warburg2020mapillary} and Pittsburgh30k~\cite{torii2013visual}. While these compact representations with low memory and low latency enable efficient large-scale retrieval tasks, they often exhibit limited robustness in challenging environments. This stems from the limited generalization of ImageNet-pretrained backbones, which are biased toward object-centric distributions, and the inadequate diversity of training datasets, which fail to capture the full spectrum of real-world variations in conditional changes, viewpoints, and scene structures. A promising solution involves two-stage methods~\cite{cao2020unifying, hausler2021patch, wang2022transvpr, zhu2023r2former, jiang2024robust}, where global features are initially used to retrieve the top-k candidates from the database, followed by re-ranking these candidates via local features matching. However, the substantial memory usage and high latency associated with local features matching pose significant challenges for practical applications, highlighting the need for robust, lightweight global descriptors.

Recent advances~\cite{lu2024towards, lu2024cricavpr, izquierdo2024optimal} leverage the capabilities of DINOv2~\cite{oquab2023DINOv2}, a transformer-based model trained on the LVD-142M dataset. These methods utilize DINOv2 as a backbone to extarct features, as an alternative to commonly used CNNs and ViTs trained on ImageNet, achieving robust, high-performance results. In parallel, CosPlace~\cite{berton2022rethinking} and EigenPlaces~\cite{berton2023eigenplaces} introduces a large-scale, VPR dense training dataset, referred to as SF-XL~\cite{chen2011city}, which eliminates the need for mining negative examples during the training process and enables robust learning from extensive data collections. However, their grid-based partitioning strategies yield inefficient clustering, causing data imbalance across classes and reducing effective utilization.
Recent advances~\cite{lu2024towards, lu2024cricavpr, izquierdo2024optimal} leverage the capabilities of DINOv2~\cite{oquab2023DINOv2}, a transformer-based model trained on the LVD-142M dataset. These methods utilize DINOv2 as a backbone to extarct features, as an alternative to commonly used CNNs and ViTs trained on ImageNet, achieving robust, high-performance results. In parallel, CosPlace~\cite{berton2022rethinking} and EigenPlaces~\cite{berton2023eigenplaces} introduces a large-scale, VPR dense training dataset, referred to as SF-XL~\cite{chen2011city}, which eliminates the need for mining negative examples during the training process and enables robust learning from extensive data collections. However, their grid-based partitioning strategies yield inefficient clustering, causing data imbalance across classes and reducing effective utilization.

\begin{figure}[t]
    \centering
    \includegraphics[width=\linewidth]{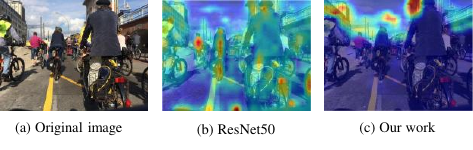}
    \caption{Feature heatmap visualizations of both the pre-trained foundation model (ResNet-50) and our approach. The pre-trained model emphasizes irrelevant regions, such as dynamic pedestrians, while our method targets discriminative areas, such as buildings, which are more relevant for place identification.}
    \label{fig:2}
\end{figure}

In this study, we propose Dual-Scale-Former (DSFormer), an novel Transformer-based model designed to integrate dual-scale features from the final two CNN layers. Unlike prior approaches~\cite{wang2022transvpr, jiang2024robust} that concatenate multi-layer features with minimal cross-layer interaction, DSFormer employs self-attention to capture long-range dependencies within each scale and a cross-attention module to dynamically fuse cross-scale correlations and allocate weights, yielding a robust global embedding attuned to semantic and structural cues, as illustrated in Fig.~\ref{fig:2}. Concurrently, we introduce a block clustering approach based on HDBSCAN\cite{mcinnes2017hdbscan} to repartition SF-XL. This partitioning approach mitigates the significant imbalance in data distribution across classes and reduces redundancy, thereby enhancing the overall efficiency of data utilization. Our contributions could be summarized as follows: 

\begin{itemize}[topsep=0.1cm, itemsep=0.1cm, parsep=0.1cm]
    \item We propose DSFormer, a Transformer-based module that leverages dual-scale feature cross-learning to produce discriminative global descriptors, enhancing robustness to environmental and viewpoint variations in VPR.
    
    \item We propose a block clustering strategy for SF-XL, optimizing data utilization efficiency and reducing volume by approximately 30\% compared to EigenPlaces, while achieving superior performance.
\end{itemize}

\section{Related work}
\label{sec:2}
Traditional VPR methods~\cite{galvez2012bags, jegou2010aggregating} apredominantly rely on aggregation-based techniques that summarize statistical properties of hand-crafted features into compact representations. VPR tasks can be designed to address image retrieval tasks by focusing solely on image content, leading to the emergence of many high-performance learning-based solutions. With the advent of deep learning, modern VPR approaches have shifted toward leveraging pretrained CNNs and ViTs backbones, integrated with advanced aggregation strategies to produce robust global descriptors. Three prominent based-learning aggregated methods NetVLAD~\cite{arandjelovic2016netvlad}, GeM~\cite{radenovic2018fine}, and MixVPR~\cite{Ali-bey_2023_WACV} exemplify this trend, each offering distinct mechanisms to enhance feature representation. NetVLAD, a pioneering approach, replaces the traditional hand-crafted clustering of VLAD with a learnable clustering layer, enabling end-to-end training and adaptable feature grouping. Its flexibility and strong performance have inspired numerous variants, such as CRN~\cite{kim2017learned}, which refines contextual relationships, TransVLAD~\cite{xu2023transvlad}, which incorporates Transformer-based enhancements, and MultiRes-NetVLAD~\cite{khaliq2022multires}, which integrates multi-resolution features to improve robustness across scales. In contrast, GeM introduces a learnable parameter to dynamically adjust sensitivity to feature magnitudes. This adaptability has made it a widely adopted baseline, with recent frameworks like CosPlace~\cite{berton2022rethinking} and EigenPlaces~\cite{berton2023eigenplaces} harnessing GeM to achieve significant performance gains in large-scale VPR by optimizing training on dense datasets like SF-XL. More recently, MixVPR~\cite{Ali-bey_2023_WACV} employ multi-layer perceptrons (MLPs) to mix deep features, producing highly discriminative global descriptors that capture complex patterns. Although these learning-based aggregated methods excel in computational efficiency and perform reliably in controlled or simple scenarios, their limitations become apparent in more challenging real-world environments.

To mitigate the limitations of global retrieval methods in challenging environments, two-stage approaches have emerged to refine VPR performance by reranking the top-k candidates retrieved in the initial global stage. These methods introduce a subsequent reranking phase that leverages spatial consistency matching of local features to enhance retrieval accuracy. Broadly, two-stage techniques can be classified into RANSAC-based and learning-based paradigms, each addressing geometric verification through distinct strategies.  RANSAC-based methods, such as Patch-NetVLAD~\cite{hausler2021patch}, DELG~\cite{cao2020unifying}, TransVPR~\cite{wang2022transvpr}, and EUPN~\cite{jiang2024robust}, estimate homography between query and reference images to identify matching inliers among local features. Candidates are then reranked based on the number of inliers, effectively incorporating spatial relationships to improve robustness against viewpoint shifts and environmental variations. Patch-NetVLAD refines NetVLAD descriptors with patch-level matching, while DELG combines global and local features in a unified framework, demonstrating versatility across diverse datasets. In contrast, learning-based approaches, such as R$^2$Former~\cite{zhu2023r2former} employs a Transformer-based matching network to replace the computationally intensive RANSAC process, ‌by‌ directly computing matching scores between local features, ‌while‌ leveraging attention mechanisms to model complex spatial dependencies, ‌and‌ streamline reranking. By incorporating geometric information from local features, two-stage methods significantly outperform standalone global retrieval techniques, particularly in scenarios with severe illumination changes, occlusions, and perceptual aliasing. However, two-stage methods suffer from high memory overhead due to local feature storage and time delays from geometric verification, limiting their practicality.

Despite advancements in retrieval accuracy through advanced reranking techniques, the performance of VPR methods depends on both the pretrained backbone and the training datasets employed. A significant leap in backbone generalization has been achieved with DINOv2~\cite{oquab2023DINOv2}, a Transformer-based model pretrained on the expansive LVD-142M dataset. This large-scale training endows DINOv2 with robust and generalizable features, surpassing the capabilities of traditional CNNs and ViTs trained on ImageNet’s object-centric distributions. AnyLoc~\cite{keetha2023anyloc} first capitalizes on this by utilizing a pretrained DINOv2 backbone to extract multi-layer features, pooled via VLAD, enabling effective cross-domain VPR with minimal reliance on task-specific annotations. 
Subsequent DINOv2-based VPR methods~\cite{lu2024towards, lu2024cricavpr, izquierdo2024optimal} further demonstrate that, this backbone achieves state-of-the-art performance in challenging scenarios (extreme illumination changes, viewpoint shifts) owing to its inherent generalization derived from DINOv2. For conventional training datasets, such as MSLS~\cite{warburg2020mapillary} and Pittsburgh30k~\cite{torii2013visual}, offer limited variability in environmental conditions, such as illumination, weather, seasonal shifts and viewpoints, resulting in models that struggle to generalize to diverse, unseen test domains. To overcome this, CosPlace~\cite{berton2022rethinking} introduces the processed SF-XL dataset, a large-scale, densely sampled collection designed to enhance robustness through extensive real-world coverage without negative mining. EigenPlaces~\cite{berton2023eigenplaces} builds upon this by reprocessing SF-XL to better handle viewpoint variations through refined partitioning. However, these efforts are hampered by suboptimal data classification, where fixed grid divisions misalign geographically proximate locations, resulting in inefficient data utilization.

\begin{figure*}[t]
\centering
    \includegraphics[width=\linewidth]{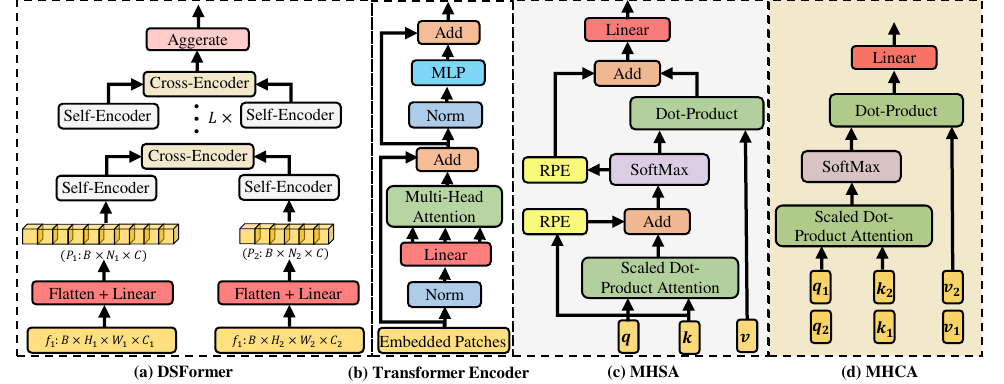}
    \caption{(a) DSFormer module: The inputs consists of dual-scale feature maps from ResNet-50, linearly mapped into two embedded patches of the same dimension, which are processed through $L$ encoding blocks with self-encoder and shared cross-encoder. (b) Transformer encoder layer. (c) Self-Encoder block employs Multi-Head Self-Attention (MHSA) with Relative Position Encoding. (d) Cross-Encoder utilizes Multi-Head Cross-Attention (MHCA) and operates on dual-scale patch embeddings as inputs.}
    \label{fig:3}
\end{figure*}

\section{Methodology}
\label{sec:3}

To tackle the inherent challenges in VPR tasks, we propose two synergistic strategies designed to enhance both feature representation and data utilization efficiency. First, we introduce DSFormer, a novel Transformer-based architecture that effectively captures and integrates dual-scale features, significantly improving the discriminability of feature representations. Second, we develop an advanced block clustering technique leveraging HDBSCAN~\cite{mcinnes2017hdbscan} to preprocess the densely SF-XL dataset~\cite{chen2011city}, optimizing the utilization efficiency of training data by ensuring coherent class assignments for geographically proximate locations.

\subsection{DSFormer}
\label{sec:3-1}

Our model still adopts ResNet-50 as the backbone for feature extraction, leveraging it to encode rich visual information. Unlike prior approaches that depend exclusively on features from either the final or penultimate layer, we harness dual-scale features extracted from the last two layers to construct more robust global descriptors. While ResNet-50 provides strong local inductive bias, it can suffer from limited global contextual awareness. To address this, we introduce DSFormer, which enables cross-learning between both layers through a synergistic combination of self-attention and cross-attention mechanisms, effectively capturing long-range dependencies and inter-scale correlations. Finally, a GeM pooling module is applied to aggregate the dual-scale features into a compact global descriptor. The complete framework, illustrating this process, is presented in Fig.~\ref{fig:3}.

\subsubsection{Transformer Encoder}
\label{sec:3-1-1}

The Vision Transformer, proposed by Dosovitskiy et al.~\cite{dosovitskiy2021an}, redefines visual processing by treating images as sequences of embeddings suitable for Transformer architectures, and in our adaptation, we replace raw image patches with feature patches derived from a pretrained backbone. Given feature maps of different scales $f_i \in \mathbb{R}^{C_i \times H_i \times W_i}, i=1, 2$, we partition them into $N_i = H_i \times W_i$ pacthes. $f_i$ is flattened into embeddings:

\begin{equation}\label{eq:1}
z_{0}=[x_{f}^{1}; x_{f}^{2}; \cdots; x_{f}^{N}]+ E_{pos}
\end{equation}

where $x_{f}^{j}$ denotes the $j$-th feature patch of $f_i$, $E_{pos}$ provides positional encodings to preserve spatial relationships. These embeddings are subsequently processed through Transformer encoding layers utilizing Multi-Head Attention (MHA):

\begin{equation}\label{eq:2}
 z_{t} =\textrm{MHA}(\textrm{LN}(z_{l-1}))+z_{l-1},
 z_{l} =\textrm{FFN}(\textrm{LN}( z_{t}))+ z_{t}
\end{equation}

\begin{equation}\label{eq:3}
\textrm{MHA}(\bm{q, k, v}) = \textrm{Concat}(\textrm{head}_1,\cdots,\textrm{head}_h)W^{O} 
\end{equation}

\begin{equation}\label{eq:4}
\textrm{head}_i = \textrm{Attention}(\bm{q}W^{\bm{q}}_{i}, \bm{k}W^{\bm{k}}_{i}, \bm{v}W^{\bm{v}}_{i}) 
\end{equation}

where $\textrm{Attention}(\bm{q}, \bm{k}, \bm{v}) = \textrm{softmax}(\frac{\bm{q}\cdot\bm{k}^{T}}{\sqrt{d_k}})\bm{v}$, This patch-based Transformer framework, designed to process feature inputs without a classification token, underpins our DSFormer, which employs these principles to integrate dual-scale features for VPR tasks.

\subsubsection{MHSA}
\label{sec:3-1-2}

To effectively model spatial relationships within feature patch sequences, we employ a customized Self-Attention by incorporating the Improving Relative Position Encoding (IRPE)~\cite{wu2021rethinking} module. Three learnable functions, denoted as $\textrm{rpe}_{\bm{q}}$, $\textrm{rpe}_{\bm{k}}$, $\textrm{rpe}_{\bm{v}}$, are employed to map the relative distances to their corresponding encoding values, as follows:

\begin{equation}\label{eq:5}
\begin{aligned}
\textrm{Attention}(\bm{q}, \bm{k}, \bm{v}) &= \textrm{rpe}_{\bm{v}}( \textrm{softmax}( \frac{\bm{q}\cdot\bm{k}^{T}}{\sqrt{d_k}} + \textrm{rpe}_{\bm{q}}(\bm{q}) \\
&\quad + \textrm{rpe}_{\bm{k}}(\bm{k}) ) \bm{v} )
\end{aligned}
\end{equation}

By incorporating IRPE, DSFormer effectively captures long-range dependencies and spatial coherence, thereby improving feature discriminability and enhancing robustness in VPR across diverse environmental conditions and viewpoint variations.

\begin{figure}[t]
\centering
    \includegraphics[width=\linewidth]{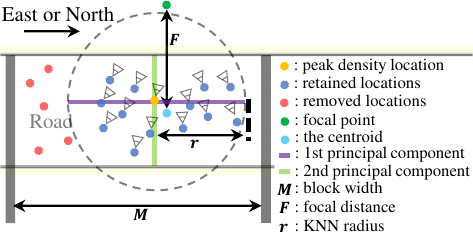}
    \caption{\textbf{Sketch of block clustering’s principle.} The region is initially partitioned along the east or north direction into blocks of width $M$ meters, where sampling locations truncated by street boundaries are grouped into a single class. The KNN method retains sampling locations within a radius $r$ centered on the peak density location, discarding others. SVD is applied to identify two principal directions, and from each centroid, a focal point is defined along these directions at a distance $F$, with images at sampling locations oriented toward this focal point.}
    \label{fig:4}
\end{figure}

\subsubsection{MHCA}
\label{sec:3-1-3}

To facilitate effective cross-learning between dual-scale features in DSFormer, we introduce a specialized shared Cross-Attention module that integrates features from the last two layers of the ResNet-50 backbone, enabling the fusion of coarse-grained and fine-grained representations, thereby improving the robustness and discriminability of global descriptors for visual place recognition tasks. Given sets of feature patch embeddings: $\bm{z_i} \in \mathbb{R}^{N_i \times C}$. The Cross-Attention is defined as:

\begin{equation}\label{eq:6}
\begin{aligned}
&\textrm{Attention}(\bm{q}_i, \bm{k}_m, \bm{v}_m) = \textrm{softmax}\left(\frac{\bm{q}_i\cdot{\bm{k}_m}^{T}}{\sqrt{d_{\bm{k}_m}}}\right)\bm{v}_m \\
&\bm{q}_i, \bm{k}_i, \bm{v}_i = \bm{z}_i(W^{\bm{q}_i}, W^{\bm{k}_i}, W^{\bm{v}_i}), i = 1, 2.
\end{aligned}
\end{equation}

\noindent where $i \neq m$ for Attention.

\subsection{Block Clustering}
\label{sec:3-2}
To improve data utilization efficiency, we introduce a block clustering approach for the repartition of the SF-XL dataset, using UTM coordinates $\{east, north\}$. The same procedure is implemented for coordinate values in both directions, with the east serving as a representative example. 

Given a UTM coordinate $x_{i, j} = (x_{e}^{i}, x_{n}^{i})$, representing the $i$-th location assigned to the $j$-th group $\mathcal{G}_{e}^{j}$, where $(x_{e}^{i}, x_{n}^{i})$ denote the easting and northing coordinates, respectively. And,

\begin{equation}\label{eq:7}
j = \left\lfloor \frac{(x_{e}^{i} - \min(\{x_{e}^{i}\}_{i=1}^{I})) \mod (M \cdot N)}{M} \right\rfloor + 1
\end{equation}

\noindent Here, $M$ is the width of each block, $N$ denotes the total number of groups, $I$ is the total number of locations. Subsequently, the group $\mathcal{G}_{e}^{j}$ is clustered into $K$ classes by HDBSCAN~\cite{mcinnes2017hdbscan}, leveraging its hierarchical approach to identify density-based clusters of varying shapes and sizes with robustness to noise. However, the uncertainty in intra-class distances can lead to sampling locations within the same class being widely separated, thereby reducing the model’s sensitivity to spatial proximity. To address this, we employ the
$k$-Nearest Neighbors (KNN) method to retain only those sampling points within a radius $r$ meters of peak density location, as shown in Fig.~\ref{fig:4}. This effectively constrains intra-class distances and mitigates data redundancy, enhancing the quality of the resulting dataset. Simultaneously, to prevent overlapping visible areas across classes, we exclude classes with peak density location less than $l$ meters apart, further reducing the size of dataset. For the north direction, we perform the same operation using the northing coordinate $x_{n}^{j}$ to obtain the set of groups $\{\mathcal{G}_{n}^{j}\}_{j=1}^{N}$. Further merging $\{\mathcal{G}_{e}^{j}\}_{j=1}^{N}$ and $\{\mathcal{G}_{n}^{j}\}_{j=1}^{N}$:

\begin{equation}\label{eq:8}
\{\mathcal{G}_{j}\}_{j=1}^{N} = \{\mathcal{G}_{e}^{j} \cup \mathcal{G}_{n}^{j}\}_{j=1}^{N}
\end{equation}

To prevent overlapping regions from occurring between distinct classes within the merged clusters, any classes with peak density location distance falls below a threshold of $l$ meters are removed again. Our clustering ensures that sampling locations along the same street are evenly distributed within their respective classes, preventing their exclusion due to insufficient intra-class samples from uneven partitioning, thus maintaining more effective classes.  


To improve robustness against viewpoint variations, we implement an image selection strategy based on the approach proposed in EigenPlaces~\cite{berton2023eigenplaces}. For each retained class, represented by its UTM coordinates $X_{k} \in \mathbb{R}^{n \times 2}$, where $n$ is the number of sampling locations in the $k$-th class, we perform singular value decomposition (SVD) to extract two principal directions for the centered matrix $\widehat{X_{k}} = X_{k} - E[X_{k}]$: one typically aligned with the road along which the vehicle travels in the captured images, and the other oriented perpendicular to it, corresponding to the roadside. We then calculate the corresponding focal point $c_{k}$ for each direction as follows:

\begin{equation}\label{eq:9}
c_{k} = E[X_{k}] + F \cdot V
\end{equation}

where $F$ denotes the focal distance, $V$ represents the orientation vector derived from SVD and $E[X_{k}]$ is the UTM coordinates of the centroid. Each group is divided into two groups based on two directions, resulting in $2N$ groups.

\section{EVALUATION}
\label{sec:4} 
In this section, we evaluate the performance of proposed DSFormer. The evaluation is structured into three key components: (1) Experiments Settings, which outlines the implementation details, datasets, evaluated metrics and compared methods; (2) Results and Analysis, which presents quantitative outcomes and comparative insights against state-of-the-art methods; and (3) Ablations, which investigates the contributions of individual components within our methodology. 

\begin{table*}[t]
  \centering
  \caption{Performance comparison of DSFormer and state-of-the-art VPR methods across multiple datasets in terms of Recall@1 (R@1) and Recall@5 (R@5). $\dagger$ represents training in our new SFXL partition.}
  \label{tab:1}
  \begin{tabular}{l l c c c c c c c c c c c c}
    \toprule
    \multirow{2}{*}{Method} & \multirow{2}{*}{Training Set} & \multicolumn{2}{c}{MSLS Val} & \multicolumn{2}{c}{MSLS Chal} & \multicolumn{2}{c}{Pitts30k} & \multicolumn{2}{c}{Tokyo24/7} & \multicolumn{2}{c}{Nordland} & \multicolumn{2}{c}{Average}\\
    \cmidrule(lr){3-4} \cmidrule(lr){5-6} \cmidrule(lr){7-8} \cmidrule(lr){9-10} \cmidrule(lr){11-12} \cmidrule(lr){13-14}
    & & R@1 & R@5 & R@1 & R@5 & R@1 & R@5 & R@1 & R@5 & R@1 & R@5 & R@1 & R@5\\
    \midrule
    \textit{Two-Stage Methods} & & & & & & & & & & & \\
    DELG~\cite{cao2020unifying} & GLD~\cite{noh2017large} & 84.5 & 90.4 & 53.8 & 62.5 & 89.6 & 95.3 & 87.6 & 93.3 & 67.9 & 77.8 & 76.7 & 83.9\\
    Patch-NetVLAD~\cite{hausler2021patch}   & MSLS & 79.5 & 86.2 & 48.1 & 57.6 & 88.7 & 94.5 & 85.1 & 87.0 & 62.6 & 70.8 & 72.8 & 79.2\\
    SP-SuperGLUE~\cite{detone2018superpoint,sarlin2020superglue} & SS~\cite{detone2018superpoint} & 84.5 & 88.8 & 50.6 & 56.9 & 88.7 & 95.1 & 88.3 & 89.2 & 65.3 & 70.4 & 75.5 & 80.1\\
    TransVPR~\cite{wang2022transvpr} & MSLS & 86.5 & 91.1 & 63.9 & 73.2 & 89.0 & 94.9 & 77.8 & 81.6 & 73.4 & 81.4 & 78.1 & 84.4\\
    R\(^2\)Former~\cite{zhu2023r2former} & MSLS & \textbf{89.7} & \textbf{95.0} & \textbf{73.0} & \textbf{85.9} & 91.1 & 95.2 & 86.3 & 90.8 & 77.1 & 89.0 & 83.3 & \textbf{91.2}\\
    \midrule
    \multicolumn{2}{l}{\textit{Retrieval Baselines on a \textbf{ResNet-50} backbone}} & & & & & & & & & & \\
    NetVLAD~\cite{arandjelovic2016netvlad} & MSLS & 78.2 & 87.1 & 55.8 & 71.5 & 74.8 & 88.4 & 48.9 & 67.3 & 53.0 & 75.4 & 62.1 & 77.9\\
    GeM~\cite{radenovic2018fine} & MSLS & 75.9 & 88.9 & 53.6 & 70.5 & 77.1 & 90.6 & 39.7 & 63.5 & 43.9 & 68.8 & 58.0 & 76.5\\
    CosPlace~\cite{berton2022rethinking} & SF-XL & 85.7 & 91.2 & 63.0 & 74.6 & 90.4 & 95.4 & 80.0 & 88.6 & 68.3 & 86.7 & 77.5 & 87.3\\
    EigenPlaces~\cite{berton2023eigenplaces} & SF-XL & 85.9 & 91.1 & 62.6 & 74.8 & 91.0 & 95.7 & 78.1 & 88.9 & 71.9 & 88.9 & 77.9 & 87.1\\
    MixVPR~\cite{Ali-bey_2023_WACV} & GSV-Cities~\cite{ali2022gsv} & 83.6 & 91.5 & 59.2 & 72.3 & 90.6 & 95.5 & 80.0 & 89.5 & 70.8 & 86.9 & 76.8 & 87.1\\
    BoQ + PCA~\cite{Ali-bey_2024_CVPR} & GSV-Cities & 85.8 & 90.8 & 63.2 & 74.2 & 90.2 & 94.7 & 80.3 & 88.9 & 73.6 & 89.5 & 78.6 & 87.6\\
    \midrule
    \textbf{GeM$^\dagger$ (ResNet-50)} & SF-XL & 87.6 & 91.9 & 65.3 & 76.0 & \textbf{91.9} & 96.0 & 82.9 & 92.1 & 75.5 & 91.1 & 80.6 & 89.4\\
    \textbf{DSFormer (ResNet-50)} & SF-XL & 88.9 & 93.2 & 68.1 & 77.6 & \textbf{91.9} & \textbf{96.4} & \textbf{88.6} & \textbf{93.7} & \textbf{81.5} & \textbf{94.1} & \textbf{83.8} & 91.0\\
    \bottomrule
    \toprule
    \multicolumn{2}{l}{\textit{Retrieval Baselines on a \textbf{DINOv2} backbone}} & & & & & & & & & & \\
    CricaVPR + PCA~\cite{lu2024cricavpr} & GSV-Cities & 86.9 & 93.6 & 69.3 & 81.9 & 92.3 & 96.8 & 87.6 & 94.9 & 88.1 & \textbf{97.5} & 84.8 & 92.9 \\
    SALAD + PCA~\cite{izquierdo2024optimal} & GSV-Cities & 90.3 & \textbf{96.2} & 73.2 & 87.0 & 91.2 & 95.8 & 92.4 & 97.1 & 84.5 & 95.7 & 86.3 & 94.4\\
    BoQ + PCA & GSV-Cities & 89.6 & 95.0 & 73.8 & 84.7 & \textbf{94.0} & \textbf{98.6} & 89.2 & 94.0 & 84.8 & 97.0 & 86.3 & 93.9\\
    \midrule
    \textbf{GeM$^\dagger$ (DINOv2)} & SF-XL & 90.3 & 95.9 & 74.6 & 87.3 & 93.5 & 97.0 & \textbf{95.9} & \textbf{97.8} & 87.7 & 97.4 & 88.4 & \textbf{95.1}\\
    \textbf{DSFormer (DINOv2)} & SF-XL & \textbf{92.7} & \textbf{96.2} & \textbf{75.8} & \textbf{87.4} & 93.6 & 97.0 & \textbf{95.9} & \textbf{97.8} & \textbf{88.4} & 97.2 & \textbf{89.3} & \textbf{95.1}\\
    \bottomrule
  \end{tabular}
\end{table*}

\subsection{Experiments Settings}
\label{sec:4-1}

\subsubsection{Implementation Details}
\label{sec:4-1-1}

Our model leverages a pre-trained ResNet-50 backbone~\cite{he2016deep} for feature extraction, truncated at the final two layers to preserve semantically rich local feature representations. All residual blocks except the last two are frozen, with linear projections applied to standardize channel dimensions for compatibility with the cross-attention mechanism. The implementation is built using PyTorch. Input images are resized to $320\times320$ with random scaling and augmented with color jittering. We train our model on the processed large-scale SF-XL dataset~\cite{chen2011city}, comprising approximately 4.2 million images across 10 groups, which captures a variety of extreme conditions. Training follows a framework akin to EigenPlaces~\cite{Berton_2023_ICCV}, employing the Large Margin Cosine Loss (LMCL)\cite{wang2018cosface} as the loss function. The model is optimized using the Adam optimizer\cite{adam} with a learning rate of $1\times10^{-5}$, while the classifier uses a separate Adam optimizer with a learning rate of 0.01. 
Training spans up to 40 epochs on an NVIDIA RTX 4080 GPU with 16 GB of memory, with a batch size of 32 images and each epoch defined as 10000 iterations over a group. The DSFormer module consists of 3 layers, with attention mechanisms utilizing 16 shared-parameter heads. For block clustering, we set the block width $M = 10$ meters, the number of groups $N=5$ (totaling 10 groups across directions), the KNN radius $r=7.5$ meters, the minimum distances $l=40$ meters between peak density locations, and the focal distance $F=15$ meters. Furthermore, we also assess DSFormer equipped with a DINOv2 (ViT-B) backbone. This model is trained using input images with a resolution of 322×322. During training, the final two layers of DINOv2 are fine-tuned. The architecture only incorporates a single DSFormer layer, which employs 12 attention heads with shared parameters.

\subsubsection{Datasets and Metric}
\label{sec:4-1-2}

We evaluate our method across five widely adopted benchmark VPR datasets and two large-scale datasets: the MSLS validation set~\cite{warburg2020mapillary}, MSLS challenge set~\cite{warburg2020mapillary}, Pittsburgh30k~\cite{torii2013visual}, Tokyo24/7~\cite{torii201524}, Nordland~\cite{sunderhauf2013we} (using the winter-summer partition from~\cite{olid2018single} as query and reference sets, respectively), SF-XL test datasets~\cite{berton2022rethinking, barbarani2023local}. For Nordland, performance is measured with an error tolerance of $\pm 2$ frames. All other datasets employ Recall@N (R@N) as the metric, where success requires at least one of the top-$N$ retrieved results to match the ground truth within a position error tolerance of 25 meters.

\begin{table}[t]
  \centering
  \small 
  \caption{Comparison of DSFormer with two-stage VPR methods in terms of latency and memory usage, evaluated on the MSLS validation set using an NVIDIA RTX 3060.}
  \label{tab:2}
  \resizebox{1\linewidth}{!}{
  \renewcommand{\arraystretch}{1.2}
  \begin{tabular}{l c c c c}
    \toprule
    \multirow{2}{*}{Method} & \multirow{2}{*}{\makecell[c]{Des. Size}} & \multirow{2}{*}{\makecell[c]{Memory \\ (GB)} } & \multicolumn{2}{c}{Latency(s)/query}  \\
    \cmidrule(lr){4-5}
    & &  & Extract & Rerank \\
    \midrule
    DELG & $1000\times128 + 2048$ & 7.8 & 0.096 & 35.8 \\
    Patch-NetVLAD & $2826\times4096+4096$ & 908.3 & 1.006 & 16.88 \\
    SP-SuperGLUE & $2048\times256+4096$ & 41.6 & 0.092 & 12.381 \\
    TransVPR & $1200\times256+256$ & 24.2 & 0.017 & 2.118 \\
    R\(^2\)Former & $500\times131+256$ & 5.2 & 0.025 & 0.423 \\
    \midrule
    \textbf{DSFormer (Ours)} & \textbf{$1\times512$} & \textbf{0.039} &\textbf{ 0.024} & \textbf{N/A} \\
    \bottomrule
  \end{tabular}}
\end{table}

\subsubsection{Compared Methods}
\label{sec:4-1-3}
We compare our DSFormer against several state-of-the-art (SOTA) VPR methods, categorized into two types. Retrieval baselines include NetVLAD~\cite{arandjelovic2016netvlad}, GeM~\cite{radenovic2018fine}, CosPlace~\cite{berton2022rethinking}, EigenPlaces~\cite{berton2023eigenplaces}, MixVPR~\cite{Ali-bey_2023_WACV}, and BoQ~\cite{Ali-bey_2024_CVPR}. For fair comparison, these methods use an input image size of $320 \times 320$ and employ ResNet-50 as the backbone network, producing global descriptors with a dimension of 512 (PCA is applied for dimensionality reduction for BoQ). Additionally, we include several recent retrieval methods that adopt the DINOv2 (ViT-B) backbone—namely CricaVPR~\cite{lu2024cricavpr}, SALAD~\cite{izquierdo2024optimal}, and BoQ. These methods use an input image resolution of $322 \times 322$, and the descriptors are reduced to 512 dimensions via PCA, ensuring a fair comparison. To maintain consistency across evaluations, we re-implemented most of the retrieval methods to ensure that the input image resolution remains the same during training and testing. Two-stage methods comprise DELG~\cite{cao2020unifying}, Patch-NetVLAD~\cite{hausler2021patch}, TransVPR~\cite{wang2022transvpr}, and R$^2$Former~\cite{zhu2023r2former}, which first leverage global descriptors to retrieve the top-100 candidates, followed by local descriptor-based reranking, and the input image size is $640 \times 480$. All comparison results reported here are from our own reproductions of the respective methods.

\subsection{Results and Analysis}
\label{sec:4-2}

We evaluate DSFormer against state-of-the-art VPR methods on standard benchmark datasets, as presented in Table~\ref{tab:1}. DSFormer (ResNet-50) achieves superior performance on average metrics, surpassing EigenPlaces by 5.9\% and R\(^2\)Former by 0.5\% in terms of R@1 on average when averaged across the five datasets. Although DSFormer’s performance on MSLS is marginally lower than that of R\(^2\)Former, it demonstrates advantages on the remaining datasets. It is noteworthy that R2Former's training dataset originates from the same source as MSLS's evaluation set, and while R2Former employs a higher input resolution of $480\times640$ compared to our model's $320\times320$ configuration, these may result in marginally inferior evaluation performance for our approach compared with R2Former's results in MSLS. Furthermore, we compare the computational cost of our method with that of two-stage methods. DSFormer achieves a memory footprint that is 132 times smaller than that of R2Former while also demonstrating substantially higher computational efficiency compared to all reranking-based approaches, as it eliminates the necessity for a reranking process, as shown in Tab.~\ref{tab:2}. DSFormer (DINOv2) achieves the best performance among DINOv2-based methods, outperforming SALAD by 3.0\% in average R@1. This result confirms the effectiveness of DSFormer even when applied with a DINOv2 backbone.

We evaluate GeM$^\dagger$ (ResNet-50) model, trained on the SF-XL dataset using our block clustering strategy, against CosPlace and EigenPlaces—both employing GeM models trained on different partitions of SF-XL, outperforming them by 3.1\% and 2.7\% in an average of R@1, respectively. This improvement is attributed to our block clustering method, which not only enhances retrieval accuracy but also reduces dataset size by approximately 25\% (4.2M vs. 5.6M) and 32\% (4.2M vs. 6.2M) compared to the partitioning techniques utilized in CosPlace and EigenPlaces, respectively. Additionally, DSFormer surpasses GeM$^\dagger$ by 3.2\% in average R@1 (ResNet-50 backbone), demonstrating the effectiveness of the DSFormer module. This advantage is particularly evident on the challenging Tokyo24/7 and Nordland datasets, where DSFormer achieves performance gains of 5.7\% and 6.0\%, respectively, further validating its robustness in handling severe appearance changes. Moreover, GeM$^\dagger$ with DINOv2 as the backbone outperforms its ResNet-50 by 7.8\% in average R@1, indicating that DINOv2 (ViT-B) possesses significantly stronger generalization capability.

\begin{table}[t]
  \centering
  \caption{Supplementary performance comparison of DSFormer and state-of-the-art VPR methods on large-scale SF-XL test sets in terms of Recall@1.}
  \label{tab:3}
  \resizebox{0.48\textwidth}{!}{
  \begin{tabular}{l c c c c c}
    \toprule
    Method & v1 & v2 & Night & Occlusion & Average\\
    \midrule
    DELG & 85.0 & \textbf{93.5} & 28.3 & 30.3 & 59.3\\
    Patch-NetVLAD & 37.9 & 77.4 & 12.2 & 13.2 & 35.2\\
    SP-SuperGLUE & 41.6 & 78.9 & 12.2 & 13.2 & 36.5\\
    TransVPR & 37.0 & 68.4 & 4.9 & 15.8 & 31.5\\
    R\(^2\)Former & 45.5 & 74.9 & 11.6 & 22.4 & 38.6 \\
    \midrule
    NetVLAD & 19.7 & 41.6 & 3.2 & 6.6 & 17.8\\
    GeM & 17.6 & 38.3 & 3.2 & 9.2 & 17.1\\
    CosPlace & 75.8 & 85.6 & 26.8 & 36.8 & 56.3\\
    EigenPlaces & 80.0 & 90.3 & 25.3 & 32.9 & 57.1\\
    MixVPR& 61.6 & 85.5 & 13.1 & 25.0 & 46.3\\
    BoQ + PCA & 60.0 & 82.9 & 15.0 & 25.0 & 45.7\\
    \midrule
    \textbf{GeM$^\dagger$ (ResNet-50)} & 82.1 & 89.5 & 29.2 & 38.2 & 59.8\\
    \textbf{DSFormer (ResNet-50)} & \textbf{85.7} & 91.3 & \textbf{31.5} & \textbf{42.1} & \textbf{62.7}\\
    \bottomrule
    \toprule
    CricaVPR + PCA & 77.5 & 88.8 & 31.1 & 47.4 & 61.2\\
    SALAD + PCA & 80.4 & 92.1 & 41.8 & 38.2 & 63.1\\ 
    BoQ + PCA & 73.6 & 87.3 & 32.4 & 39.5 & 58.2\\
    \midrule
    \textbf{GeM$^\dagger$ (DINOv2)} & 93.5 & 94.3 & 52.1 & 47.4 & 71.8\\
    \textbf{DSFormer (DINOv2)} & \textbf{93.7} & \textbf{94.8} & \textbf{54.1} & \textbf{52.6} & \textbf{73.8}\\
    \bottomrule
  \end{tabular}
  }
\end{table}

In real-world scenarios, databases often consist of an extensive collection of reference images, while query images may not necessarily originate from the same dataset as the reference images. To evaluate the performances of VPR methods in these challenging scenarios, we add a novel large-scale benchmark, SF-XL, which includes multiple query subsets (v1, v2, night, and occlusion). The database for this benchmark contains a substantial number of reference images (about 2.8M), whereas the query images of most query subsets are not come from SF-XL and encompass complex conditions, such as night-time environments and significant occlusions. We evaluate the performance of global retrieval methods and two-stage approaches on this benchmark, with the results summarized in Table~\ref{tab:3}. Our proposed DSFormer (ResNet-50) achieves the highest accuracy, surpassing EigenPlaces by an average of 5.6\% in R@1. The suboptimal results of two-stage approaches can be attributed to their heavy reliance on global retrieval accuracy. Moreover, our DSFormer (DINOv2) still achieves outstanding performance, surpassing SALAD by 10.7\%.

\begin{table}[t]
  \centering
  \caption{Ablation study on the number of DSFormer layers and the impact of IRPE, Self-Encoder(SE), Cross-Encoder(CE) and Block Clustering(BC), showing Recall@1 results with a 512-dimensional output across multiple datasets.}
  \label{tab:4}
  \begin{tabular}{l c c c c}
    \toprule
    Ablated Versions & MSLS Val & Pitts30K & Tokyo24/7 & Nordland \\
    \midrule
    \textbf{Three layers} &  88.9 & 91.9 & 88.6 & 81.5 \\
    \midrule
    Zero layer & 87.2 & 91.5 & 81.9 & 77.7\\
    One layer & 87.6 & 91.7 & 85.1 & 81.0\\
    Two layers & 88.6 & 92.2 & 86.7 & 80.8 \\
    Four Layers & 87.7 & 92.0 & 87.6 & 82.8 \\
    \midrule
    Remove IRPE & 88.4 & 91.9 & 88.3 & 80.8\\
    Remove SE & 87.8 & 91.6 & 87.0 & 81.9\\
    Remove CE & 87.7 & 91.9 & 86.0 & 83.0\\
    \midrule
    Remove BC & 87.4 & 91.9 & 88.3 & 77.2 \\
    \bottomrule
  \end{tabular}
\end{table}
\subsection{Ablations}
\label{sec:5-3}

\subsubsection{Ablation on DSFormer}
\label{sec:5-3-1}

In the ablation study, we systematically analyzed the influence of the number of DSFormer layers and its core components on model performance. The results, presented in Table~\ref{tab:4}, are divided into two main parts. The first section investigates the effect of varying the number of DSFormer layers. When all DSFormer layers are removed, the model exhibits a noticeable decline in performance on datasets characterized by severe appearance changes, such as Tokyo24/7 and Nordland datasets. Optimal results on MSLS Val and Tokyo24/7 are achieved when employing three DSFormer layers. However, increasing the number of layers to four does not provide substantial benefits and, in some cases, leads to a slight performance decline. Therefore, we adopt the three-layer DSFormer configuration as the baseline, as it provides a better trade-off between performance and efficiency. The second section examines the contribution of individual DSFormer components. We independently removed IRPE, the Self-Encoder(SE), and the Cross-Encoder(CE), each of which resulted in a performance degradation across multiple datasets. These findings underscore the importance of the DSFormer module in improving model robustness and overall effectiveness. The final section presents an ablation study of the proposed block clustering strategy (removing the clustering is equivalent to using the data partitioning scheme employed in EigenPlaces for the SF-XL dataset), where a performance drop is observed across multiple evaluation datasets, thereby validating the effectiveness of the clustering design.

\begin{table}[h]
  \centering
  \caption{Ablation study on focal distance, showing Recall@1 results for the GeM model with a ResNet-50 backbone and 512-dimensional output on the MSLS Val and Pitts30K datasets.}
  \label{tab:5}
  \begin{tabular}{c c c c c c c}
    \toprule
    \multirow{2}{*}{Focal Distance(m)} & \multicolumn{3}{c}{MSLS Val} & \multicolumn{3}{c}{Pitts30K} \\
    \cmidrule(lr){2-4} \cmidrule(lr){5-7}
    & R@1 & R@5 & R@10 & R@1 & R@5 & R@10\\
    \midrule
    5 & 85.7 & 91.1 & 92.4 & 91.2 & 96.1 & 97.0 \\
    10 & 87.4 & 92.0 & 93.1 & 91.5 & 96.3& 97.3 \\
    15 & 87.6 & 91.9 & 93.2 & 91.9 & 96.0 & 97.1\\
    20 & 87.4 & 91.8 & 93.2 & 91.6 & 96.2 & 97.2 \\
    25 & 87.4 & 92.6 & 93.8 & 92.0 & 96.3 & 97.3 \\
    30 & 87.3 & 92.0 & 93.6 & 91.7 & 96.0 & 97.2\\
    \bottomrule
  \end{tabular}
\end{table}

\subsubsection{Ablation on the focal distance}
\label{sec:5-3-2}
In this section, we conducted ablation experiments to investigate the effect of an important parameter in block clustering: focal distance, defined as the distance from the class centroid to the focal point. This parameter influences the orientation and perspective of sampled images, potentially affecting the model's performance in visual localization tasks. To evaluate its impact, we tested the GeM$^\dagger$ (ResNet-50) model across various focal distances on the MSLS Val and Pitts30K datasets, as shown in Table~\ref{tab:5}. At shorter distances (e.g., 5m), the limited field of view may constrain the model’s ability to capture comprehensive scene information, resulting in lower accuracy. As the focal distance increases to 15m, the wider perspective enhances the model’s understanding of the environment, boosting performance. However, excessively large distances (e.g., 30m) offer diminishing returns, as the broader perspective may introduce irrelevant details or noise, leading to minor performance fluctuations. Based on these findings, we select 15m as the default focal distance, as it balances capturing sufficient contextual information with avoiding excessive noise. 

\section{Conclusions}
This study introduces DSFormer, an advanced model designed to address fundamental challenges in Visual Place Recognition (VPR), particularly variations in viewpoint and appearance. We optimize data partitioning within the SFXL dataset by incorporating a novel block clustering strategy,  enhancing training efficiency and reduce data redundancy. Additionally, the proposed Dual-Scale Transformer (DSFormer) module improves feature representation by leveraging a dual-scale cross-learning mechanism. Experimental evaluations  are conducted on multiple benchmark datasets,  demonstrating that DSFormer effectively captures robust feature representations, enabling it to handle diverse challenges and outperform existing state-of-the-art approaches.

\bibliographystyle{IEEEtran}
\bibliography{IEEEabrv,bibfile}
\end{document}